\documentclass[11pt,a4paper]{article}
\usepackage[hyperref]{acl2020}
\usepackage{times}
\usepackage{latexsym}
\usepackage{xspace}
\usepackage{dsfont}

\usepackage{titlesec}
\usepackage{etoolbox}

\makeatletter
\patchcmd{\ttlh@hang}{\parindent\z@}{\parindent\z@\leavevmode}{}{}
\patchcmd{\ttlh@hang}{\noindent}{}{}{}
\makeatother

\titlespacing*{\section}
{0pt}{2.5ex plus 0.5ex minus 0.5ex}{1ex plus 0.2ex minus 0.2ex}
\titlespacing*{\subsection}
{0pt}{2ex plus 0.5ex minus 0.3ex}{1ex plus 0.2ex minus 0.2ex}

\aclfinalcopy %
\def\aclpaperid{2716} %

\usepackage{url}
\usepackage[utf8]{inputenc}
\usepackage{graphicx}
\usepackage{caption}
\usepackage{subcaption}

\usepackage{comment}
\usepackage{amsmath}
\usepackage{amssymb}
\usepackage{amsfonts}
 \usepackage[bottom]{footmisc}
\usepackage[parfill]{parskip}
\usepackage{hyperref}
\hypersetup{breaklinks=true,colorlinks=true}
\usepackage{todonotes}
\usepackage{color}
\usepackage{pifont}

\usepackage{array}
\usepackage{booktabs}
\usepackage{arydshln}

\newcommand\blfootnote[1]{%
  \begingroup
  \renewcommand\thefootnote{}\footnote{#1}%
  \addtocounter{footnote}{-1}%
  \endgroup
}

\newcommand\ie{\emph{i.e.}}
\newcommand\eg{\emph{e.g.}}

\newcommand{\ignore}[1]{}

\newcommand{\Table}[1]{Table~\ref{#1}}

\newcommand{\Figure}[1]{Figure~\ref{#1}}

\newcommand{\Section}[1]{Sec.~\ref{#1}}
\newcommand{\Appendix}[1]{Appendix~\ref{#1}}

\newcommand{\zhukov}[0]{\textsc{OrderedDiscrim}\xspace}
\newcommand{\richard}[0]{\textsc{ActionSets}\xspace}
\newcommand{\steprecall}[0]{step recall\xspace} %
\newcommand{\Recovery}[0]{Recall\xspace} %
\newcommand{\bkg}[0]{\textsc{BKG}\xspace} %

\newcommand{\timestep}[0]{timestep\xspace}
\newcommand{\timesteps}[0]{timesteps\xspace}
\newcommand{\Timesteps}[0]{Timesteps\xspace}
\newcommand{\Timestep}[0]{Timestep\xspace}

\newcommand{\acclabel}[0]{label\xspace} %
\newcommand{\Acclabel}[0]{Label\xspace}  %

\newcommand\an[1]{\textcolor{green}{AN: #1}}
\newcommand\df[1]{\textcolor{purple}{DF: #1}}

\title{Learning to Segment Actions from Observation and Narration}

\author{Daniel Fried$^\ddagger$ ~ Jean-Baptiste Alayrac$^\dagger$ ~ Phil Blunsom$^\dagger$ ~ \\
{\bf Chris Dyer$^\dagger$ ~ Stephen Clark$^\dagger$ ~ Aida Nematzadeh$^\dagger$ } \\
$^\dagger$DeepMind, London, UK \\
$^\ddagger$Computer Science Division, UC Berkeley ~ \\
{\tt \footnotesize dfried@berkeley.edu, \{jalayrac,pblunsom,cdyer,clarkstephen,nematzadeh\}@google.com}
}

\date{}

\begin{document}
\maketitle
\begin{abstract}
We apply a generative segmental model of task structure, guided by narration, to action segmentation in video.
We focus on unsupervised and weakly-supervised settings where no action labels are known during training.
Despite its simplicity, our model performs competitively with previous work on a dataset of naturalistic instructional videos.
Our model allows us to vary the sources of supervision used in training, and we find that both task structure and narrative language provide large benefits in segmentation quality.
\blfootnote{
Work begun while DF was interning at DeepMind. 
Code is available at \href{https://github.com/dpfried/action-segmentation}{https://github.com/dpfried/action-segmentation}.
}
\end{abstract}

\section{Learning to Segment Actions}
\label{sec:introduction}
Finding boundaries in a continuous stream %
is a crucial process for human cognition \citep{martin2003segmenting, zacks2007event,levine2019finding, unal2019event}.
To understand and remember what happens in the world around us, we need to recognize the action boundaries as they unfold and also distinguish the important actions from the insignificant ones.
This process, referred to as \emph{temporal action segmentation}, %
 is also an important first step in systems that ground natural language in videos \citep{anne2017localizing}.
These systems must identify which frames in a video depict actions -- which amounts to distinguishing these frames from background ones -- and identify which actions (\eg, boiling potatoes) each frame depicts.
Despite recent advances \citep{miech19howto100m, sun2019contrastive}, unsupervised action segmentation in videos remains a challenge.

The recent availability of large datasets of naturalistic instructional videos provides an opportunity for modeling of action segmentation  in a rich task context
\citep{yu14instructional, zhou18towards, zhukov2019cross, miech19howto100m, tang2019}; in these videos, a person teaches a specific high-level \emph{task} (\eg, making croquettes) while describing the lower-level \emph{steps} involved in that task (\eg, boiling potatoes). 
However, the real-world nature of these datasets introduces many challenges. 
For example,  more than 70\% of the frames in one of the YouTube instructional video datasets, CrossTask \citep{zhukov2019cross}, consist of \emph{background} regions (\eg, the video presenter is thanking their viewers), which do not correspond to any of the steps for the video's task.

These datasets are interesting because they provide 
(1) narrative language that roughly corresponds to the activities demonstrated in the videos and %
(2) structured task scripts that define a strong signal of the order in which steps in a task are typically performed.
As a result, these datasets provide an opportunity to study the extent to which task structure and language can guide action segmentation.
Interestingly, young children can segment actions without any explicit supervision \citep{baldwin2001infants, sharon1998individuation}, by tapping into similar cues -- action regularities and language descriptions \citep[\eg,][]{levine2019finding}.%

While previous work mostly focuses on building action segmentation models that perform well on a few metrics \citep{richard2018action, zhukov2019cross}, we aim to provide insight into how various  modeling choices impact action segmentation. %
How much do unsupervised models improve when given implicit supervision from task structure and language, and which types of supervision help most? Are discriminative or generative models better suited for the task? Does explicit structure modeling improve the quality of segmentation? 
To answer these questions, we compare two existing models with a generative hidden semi-Markov model, varying the degree of supervision. %

On a challenging and naturalistic dataset of instructional videos \cite{zhukov2019cross}, we find that our model and models from past work both benefit substantially from the weak supervision provided by task structure and narrative language, even on top of rich features from state-of-the-art pretrained action and object classifiers.
Our analysis also shows that: (1) Generative models tend to do better than discriminative models of the same or similar model class at learning the full range of step types, which benefits action segmentation; (2) Task structure affords strong, feature-agnostic baselines that are difficult for existing systems to surpass; 
(3) Reporting multiple metrics is necessary to understand each model's effectiveness for action segmentation; we can devise feature-agnostic baselines that perform well on single metrics despite producing low-quality action segments.%

\section{Related Work}

Typical methods~\citep{rohrbach2012database,singh2016multi,xu2017r,zhao2017temporal, lea2017temporal, yeung2018every, farha2019ms} for temporal action segmentation 
consist of assigning action classes to intervals of videos and rely on manually-annotated supervision.
Such annotation is difficult to obtain at scale.
As a result, recent work has focused on training such models with less supervision: one line of work assumes that only the order of actions happening in the video is given and use this weak supervision to perform action segmentation~\citep{bojanowski2014weakly,huang2016connectionist,kuehne17weakly,richard2017weakly,ding2018weakly,chang2019d3tw}.
Other approaches weaken this supervision and use only the set of actions that occur in each video \citep{richard2018action}, or are fully unsupervised \citep{sener2018unsupervised,kukleva2019unsupervised}.

Instructional videos have gained interest over the past few years~\citep{yu14instructional,Sener15unsupervised,malmaud2015s,alayrac2016unsupervised,zhukov2019cross} since they enable weakly-supervised modeling:
previous work most similar to ours consists of models that localize actions in narrated videos with minimal supervision ~\citep{alayrac2016unsupervised,Sener15unsupervised,elhamifar2019unsupervised,zhukov2019cross}.

We present a generative model of action segmentation that incorporates duration modeling, narration and ordering constraints, and can be trained in all of the above supervision conditions by maximizing the likelihood of the data; while these past works have had these individual components, they have not yet all been combined.

Our work is also related to work on inducing or predicting scripts \cite{schank2013scripts} either from text \cite{chambers2008unsupervised,pichotta2014scripts,rudinger2015learning,weber-etal-2018-hierarchical} or in a grounded setting \cite{bisk-etal-2019-benchmarking}.

\section{The CrossTask Dataset}
\label{sec:setting}

\begin{figure*}[!ht]
\centering
\includegraphics[width=\textwidth]{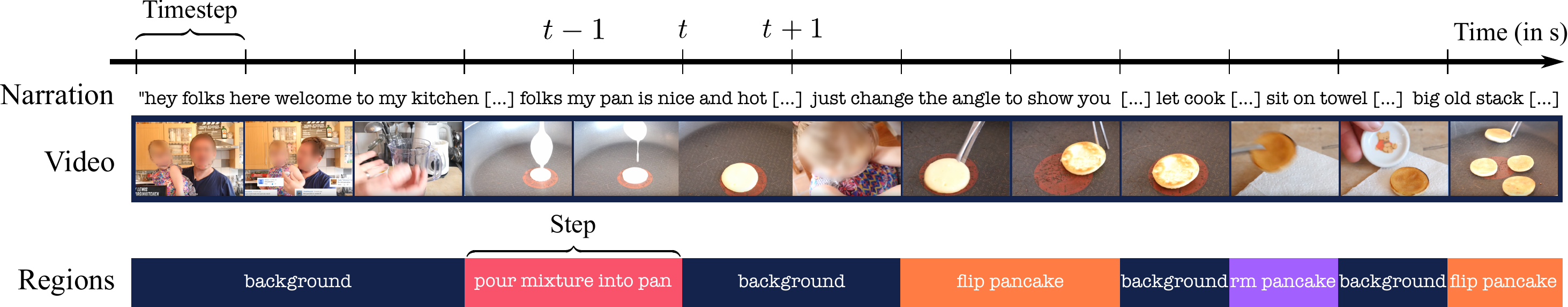}
\caption{
\label{fig:teaser}
An example video instance from the CrossTask dataset (\Section{sec:setting}). The video depicts a task, \emph{make pancakes}, and is annotated with \emph{region} segments, which can be either action steps (\eg, \emph{pour mixture into pan}) or background regions. 
Videos also are temporally-aligned with transcribed narration. 
We learn to segment the video into these regions and label them with the action steps (or background), without access to region annotations during training. 
}
\end{figure*}

We use the recent CrossTask dataset \cite{zhukov2019cross} of instructional videos. To our knowledge, CrossTask is the only available dataset that has tasks from more than one domain, includes background regions, provides step annotations and naturalistic language. Other datasets lack one of these; \eg they focus on one domain \citep{Kuehne12} or do not have natural language \citep{tang2019} or step annotations \citep{miech19howto100m}. An example instance from the dataset is shown in \Figure{fig:teaser}, and we describe each aspect below.

\paragraph{Tasks} Each video comes from a 
\emph{task},
e.g. \emph{make a latte}, with tasks taken from the titles of selected WikiHow articles, and videos curated from YouTube search results for the task name. We focus on the \emph{primary} section of the dataset, containing 2,700 videos from 18 different tasks. %

\paragraph{Steps and canonical order} Each task has a set of \emph{steps}: lower-level action \emph{step} types, \eg, \emph{steam milk} and \emph{pour milk}, which are typically completed when performing the task. Step names consist of a few words, typically naming an action and an object it is  applied to.
The dataset also provides a \emph{canonical step order} for each task: an ordering, like a script \citep{schank2013scripts, chambers2008unsupervised}, in which a task's steps are typically performed.
For each task, the set of step types and their canonical order were hand-constructed by the dataset creators based on section headers in the task's WikiHow article.

\paragraph{Annotations} Each video in the primary section of the dataset is annotated with labeled temporal segments identifying where steps occur. %
 (In the weak supervision setting, these step segment labels are used only in evaluation, and never in training.) 
A given step for a task can occur multiple times, or not at all, in any of the task's videos. Steps in a video also \emph{need not} occur in the task's canonical ordering (although in practice our results show that this ordering is a helpful inductive bias for learning).
Most of the frames in videos (72\% over the entire corpus) are \emph{background} -- not contained in any step segment.%
\paragraph{Narration} Videos also have narration text (transcribed by YouTube's automatic speech recognition system) which typically consists of a mix of the task demonstrator describing their actions and talking about unrelated topics.
Although narration is temporally aligned with the video, and steps (\eg, \emph{pour milk}) are sometimes mentioned, these mentions often do not occur at the same time as the step they describe (\eg, ``let the milk cool before \emph{pouring it}'').
\citet{zhukov2019cross} guide weakly-supervised training using the narration by defining a set of \emph{narration constraints} for each video, which identify where in the video steps are likely to occur, using similarity between the step names and temporally-aligned narration (see \Section{sec:supervision_conditions}). %

\ignore{
In the action segmentation task, we aim to segment videos into steps (or background regions which contain no steps), using features extracted from the video, using as little supervision as possible. 
Past work has guided this segmentation process using \emph{weak-supervision} in the form of the ordered list of steps.
In the weakly-supervised setting, we have access to the narration for the video, using as little supervision as possible.
We assume that we know what task a given video is from, and the step types that are possible to occur in that task (but not whether they do actually occur in the video, or what order).
}

\ignore{
    \item Reminder of why the dataset in interesting and challenging 
    \item Dataset properties and compare with other datasets
\df{while this dataset is larger than Breakfast, it's much more diverse}
\an{Note that larger datasets are available (are they?) now.}
    \item maybe an example?
    \item Analysis of the data led us to use a HSMM
}

\section{Model}
\label{sec:structured_model}

Our generative model of the video features and labeled task segments is a first-order semi-Markov model. We use a semi-Markov model for the action segmentation task because 
it explicitly models temporal regions of the video,
their duration, their probable ordering, and their features.\footnote{
Semi-Markov models are also shown to be successful in the similar domain of speech recognition (\eg, \citealp{pylkkonen2004duration}).
} 
It can be trained in an unsupervised way, without labeled regions, to maximize the likelihood of the features.

\paragraph{\Timesteps} Our atomic unit is a one-second region of the video, which we refer to as a \emph{\timestep}. A video with $T$ \timesteps has feature vectors $x_{1:T}$.
The features $x_t$ at \timestep $t$ are derived from the video, its narration, or both, and in our work (and past work on the dataset) are produced by pre-trained neural models which summarize some non-local information in the region containing each \timestep, which we describe in \Section{sec:feature_models}.

\paragraph{Regions}
Our model segments a video with $T$ \timesteps
into a sequence of regions, each of which consists of a consecutive number of \timesteps (the region's \emph{duration}). 
The number of regions $K$ in a video and the duration $d_k$ of each region can vary; the only constraint is that the sum of the durations equals the video length: $\sum_{k=1}^K d_k = T$. 
Each region has a label $r_k$, which is either one of the task's step labels (\eg, \emph{pour milk}) or a special label \bkg indicating the region is background. 
In our most general, unconstrained model, a given task step can occur multiple times (or not at all) as a region label in any video for the task, allowing step repetitions, dropping, and reordering. 

\paragraph{Structure}
We define a first-order Markov (bigram) model over these region labels:
\vspace{-.15cm}
\begin{equation}
  P(r_{1:K}) = P(r_1) \prod_{k=2}^K P(r_{k} \mid r_{k-1})
\end{equation}
with tabular conditional probabilities.
While region labels are part of the dataset, they are primarily used for evaluation: we seek models that can be trained in the unsupervised and weakly-supervised conditions where labels are unavailable.
This model structure, while simple, affords a dynamic program allowing efficient enumeration over both all possible segmentations of the video into regions and assignments of labels to the regions, allowing unsupervised training (\Section{sec:training}).

\paragraph{Duration}
Our model, following past work \cite{richard2018action}, parameterizes region durations using Poisson distributions, where each label type $r$ has its own mean duration $\lambda_r$: 
$d_k \sim \mathrm{Poisson}(\lambda_{r_k})$. 
These durations are constrained so that they partition the video: \eg, region $r_2$ begins at \timestep $d_1$ (after region $r_1$), and the final region $r_K$ ends at the final \timestep $T$.

\paragraph{\Timestep labels}
The region labels $r_{1:K}$ (step, or background) and region durations $d_{1:K}$ together give a sequence of \emph{\timestep labels} $l_{1:T}$ for all \timesteps, where a \timestep's label is equal to the label for the region it is contained in.

\paragraph{Feature distribution}
Our model's feature distribution
$p(x_{t} | l_t)$ is a class-conditioned multivariate Gaussian distribution: $x_{t} \sim \mathrm{Normal}(\mu_{l_t}, \Sigma)$,  where $l_t$ is the step label at \timestep $t$. (We note that the assignment of labels to steps is latent and unobserved during unsupervised and weakly-supervised training.)
We use a separate learned mean $\mu_{l}$ for each label type $l$, both steps and background. 
Labels are atomic and task-specific, \eg, the step type \emph{pour milk} when it occurs in the task \emph{make a latte} does not share parameters with the step \emph{add milk} when it occurs in the task \emph{make pancakes}.\footnote{
We experimented with sharing steps, or step components, across tasks in initial experiments, but found that it was helpful to have task-specific structural probabilities.
}
We use a diagonal covariance matrix $\Sigma$ which is 
fixed to the empirical covariance of each feature dimension.\footnote{We found that using a shared diagonal covariance matrix outperformed using full or unshared covariance matrices.}

\subsection{Training}
\label{sec:training}
In the unsupervised setting, labels $l$ %
are unavailable at training (used only in evaluation). We describe training in this setting, as well as two supervised training methods which we use to analyze properties of the dataset and compare model classes.

\paragraph{Unsupervised}
\label{sec:training_unsupervised_structured}
We train the generative model as a \emph{hidden} semi-Markov model (HSMM). We optimize the model's parameters to maximize the log marginal likelihood of the features for all video instance features $x^{(i)}$ in the training set: 
\vspace{-0.25cm}
\begin{align}
\mathcal{ML} = \sum_{i}^N \log P(x^{(i)}_{1:T_i})    
\label{eq:hsmm-gen}
\end{align}
 Applying the 
semi-Markov forward algorithm \cite{murphy2002hsmm,yu2010hidden}
allows us to marginalize over all possible sequences of step labels to compute the log marginal likelihood for each video as a function of the model parameters, which we optimize directly using backpropagation and mini-batched gradient descent with the Adam \cite{kingma2014adam} optimizer.\footnote{This is the same as mini-batched Expectation Maximization using gradient descent on the M-objective \cite{eisner2016inside}.}  
See \Appendix{sec:a_optimization} for optimization details.

\paragraph{Generative supervised}
Here the labels $l$ are observed; we train the model as a generative semi-Markov model (SMM) to maximize the log joint likelihood:
\vspace{-0.3cm}
\begin{align}
\mathcal{JL} = \sum_{i}^N \log P(l^{(i)}_{1:T_i}, x^{(i)}_{1:T_i})
\label{eq:smm-gen}
\end{align}
We maximize this likelihood over the entire training set using the closed form solution given the dataset's sufficient statistics (per-step feature means, average durations, and step transition frequencies). %

\paragraph{Discriminative supervised}
To train the SMM model discriminatively in the supervised setting, we use gradient descent to maximize the log conditional likelihood: %
\begin{align}
\mathcal{CL} &= \sum_{i}^N \log P(l^{(i)}_{1:T} \mid x^{(i)}_{1:T}) %
\label{eq:smm-disc}
\end{align}

\section{Benchmarks}
\begin{table}
\centering
\scalebox{0.95}{
\scriptsize{
\begin{tabular}{lcccc}
\toprule
                 & \citet{richard2018action} & \citet{zhukov2019cross} & Ours  \\
step reordering  & \checkmark       &        & \checkmark \\
step repetitions & \checkmark       &        & \checkmark \\
step duration    & \checkmark       &        & \checkmark \\
language         &              & \checkmark & \checkmark \\
generative model & \checkmark       &        & \checkmark \\
\bottomrule
\end{tabular}
}
}
\caption{
\label{tb:model-features}
Characteristics of each model we compare.
}
\vspace{-0.5cm}
\end{table}

We identify five modeling choices made in recent work: imposing a fixed  ordering on steps (not allowing step reordering); allowing for steps to repeat in a video; modeling the duration of steps; using the language (narrations) associated with the video; and using a discriminative/generative model. 
We picked the recent models of \citet{zhukov2019cross} and \citet{richard2018action} since they have non-overlapping strengths (see \Table{tb:model-features}).%

\paragraph{\zhukov} \cite{zhukov2019cross}. This work uses a discriminative classifier which gives a probability distribution over labels at each timestep: $p(l_t \mid x_t)$. %
Inference 
finds an assignment of steps to \timesteps that maximizes $\sum_{t} \log p(l_t | x_t)$ subject to the constraints that: all steps are predicted exactly once; steps occur in the fixed canonical ordering defined for the task; one background region occurs between each step. %
Unsupervised training of the model 
alternates 
between inferring labels 
using the dynamic program, and updating the classifier to maximize the probability of these inferred labels.\footnote{
To allow the model to predict step regions with duration longer than a single timestep, we
modify this classifier to also predict a background class, and incorporate the scores of the background class into the dynamic program.
}

\paragraph{\richard} \cite{richard2018action}. This work uses a generative model which has structure similar to ours, but uses dataset statistics (\eg, average video length and number of steps) to learn the structure distributions, rather than setting 
parameters to maximize the likelihood of the data. 
As in our model, region durations are modeled using a class-conditional Poisson distribution. The feature distribution is modeled using Bayesian inversion of a discriminative classifier (a multi-layer perceptron) with an estimated label prior.
The structural parameters of the model (durations and class priors) are estimated using the length of each video, and the number of possible step types.
As originally presented, this model depends on knowing which steps occur in a video at training time; for fair comparison, we adapt it to the same supervision conditions of \citet{zhukov2019cross} by enforcing the canonical step ordering for the task during both training and evaluation.

\section{Experimental Setting}
\label{sec:experiments}

We compare models on the CrossTask dataset across supervision conditions. We primarily evaluate the models on action segmentation (\Section{sec:introduction}). Past work on the dataset \cite{zhukov2019cross} has focused on a \emph{step recognition task}, where models identify individual \timesteps in videos that correspond to possible steps; for comparison, we also report performance for all models on this task. %

\subsection{Supervision Conditions}
\label{sec:supervision_conditions}
In all settings, the task for a given video is known (and hence the possible steps), but the settings vary in the availability of other sources of supervision: step labels for each \timestep in a video, and constraints from language and step ordering. Models are trained on a training set and evaluated on a separate held-out testing set, consisting of different videos (from the same tasks).

\paragraph{Supervised} Labels for all \timesteps $l_{1:T}$ are provided for all videos in the training set.  

\paragraph{Fully unsupervised} No labels for \timesteps are available during training. The only supervision is the number of possible step types for each task (and, as in all settings, which task each video is from). In evaluation, the task for a given video (and hence the possible steps, but not their ordering) are known. 
We follow past work in this setting \cite{Sener15unsupervised,sener2018unsupervised} by finding a mapping from model states to region labels that maximizes label accuracy, averaged across all videos in the task. See \Appendix{sec:a_unsupervised_evaluation} for details.

\paragraph{Weakly supervised} 
No labels for \timesteps are available, but two supervision types are used in the form of constraints~\citep{zhukov2019cross}:

(1) \emph{Step ordering constraints}: Step regions are constrained to occur in the canonical step ordering (see \Section{sec:setting}) for the task, but steps may be separated by background. 
We constrain the structure prior distributions $p(r_1)$ and transition distributions $p(r_{k+1} | r_k)$ of the HSMM to enforce this ordering.
For $p(r_1)$, we only allow non-zero probability for the background region, \bkg, and for the first step in the task's ordering. $p(r_k \mid r_{k-1})$ constrains each step type to only transition to the next step in the constrained ordering, or to \bkg.\footnote{To enforce ordering when steps are separated by \bkg, we annotate \bkg labels with the preceeding step type (but all \bkg labels for a task share feature and duration parameters, and are merged for evaluation).}
As step ordering constraints change the parameters of the model, when we use them we enforce them during both training and testing.
While this obviates most of the learned structure of the HSMM, the duration model (as well as the feature model) is still learned.

(2) \emph{Narration constraints}: %
These give regions in the video where each step type is likely to occur.
\citet{zhukov2019cross} obtained these using similarities between word vectors for the transcribed narration and the words in the step labels, and a dynamic program to produce constraint regions that maximize these similarities, subject to the step ordering matching the canonical task ordering.
See Zhukov et al.\ for details.
We enforce these constraints in the HSMM by penalizing the feature distributions to prevent 
any step labels that occur outside of one of the allowed constraint regions for that step.
Following Zhukov et al., we only use these narration constraints during training.\footnote{We also experiment with using features derived from transcribed narration in \Appendix{sec:narration_features}.}

\subsection{Evaluation}
\label{sec:metrics}

We use three metrics from past work, outlined here and described in more detail in \Appendix{sec:a_evaluation_metrics}. To evaluate action segmentation, we use two varieties of the standard \acclabel accuracy metric \cite{sener2018unsupervised,richard2018action}: 
{\bf all \acclabel accuracy}, which is computed on all \timesteps, including background and non-background, 
as well as {\bf step \acclabel accuracy}: accuracy only for \timesteps that occur in a non-background region (according to the ground-truth annotations). %
Since these two accuracy metrics  are defined on individual frames, they penalize models if they don't capture the full temporal extent of actions in their predicted segmentations.
Our third metric is {\bf \steprecall}, used by past work on the CrossTask dataset \cite{zhukov2019cross} to measure \emph{step recognition} (defined in \Section{sec:experiments}).
This metric evaluates the fraction of step types which are correctly identified by a model when it is allowed to predict only one frame per step type, per video. A high step recall indicates a model can accurately identify at least one representative frame of each action type in a video.

We also report three other statistics to analyze the predicted segmentations: 
(1) Sequence similarity: the similarity of the sequence of region labels predicted in the video to the groundtruth, using inverse Levenshtein distance normalized to be between 0 and 100. 
See \Appendix{sec:a_evaluation_metrics} for more details.   
(2) Predicted background percentage: the percentage of \timesteps for which the model predicts the background label.  Models with a higher percentage than the ground truth background percentage (72\%) are overpredicting background.
(3) Number of segments: the number of step segments predicted in a video. Values higher than the ground truth average (7.7) indicate overly-fragmented steps.
Sequence similarity and number of segments are particularly relevant for measuring the effects of structure, as they do not factor over individual timesteps (as do the all \acclabel and step \acclabel accuracies and \steprecall).

We average values 
across the 18 tasks in the evaluation set (following \citealt{zhukov2019cross}).%

\subsection{Features}
\label{sec:feature_models}
For our features $x_{1:T}$, we use the same base features as \citet{zhukov2019cross}, which are produced by convolutional networks pre-trained on separate activity, object, and audio classification datasets. See \Appendix{sec:a_feature_models} for details. 
In our generative models, we apply PCA (following \citealp{Kuehne12} and \citealp{richard2018action}) to project features to 300 dimensions and decorrelate dimensions (see \Appendix{sec:a_feature_models} for details).\footnote{This reduces the number of parameters that need to be learned in the emission distributions, both by reducing the dimensionality and allowing a diagonal covariance matrix. In early experiments we found PCA improved performance.} %

\begin{table*}[t]
\small
\centering
\scalebox{0.98}{
\begin{tabular}{clccc>{\color{gray}}c>{\color{gray}}c>{\color{gray}}c}
\toprule
   &       & All \Acclabel & Step \Acclabel & Step     & Sequence   & Predicted & Num. \\
\# & Model & Accuracy  & Accuracy   & \Recovery & Similarity & Bkg. \% & Segments. \\
\midrule
\multicolumn{5}{l}{\bf Dataset Statistic Baselines (\Section{sec:dataset_statistic_baselines})} \\
GT & Ground truth & 100.0 & 100.0 & 100.0 & 100.0 & 71.9 & 7.7 \\
B1 & Predict background &  71.9 & 0.0 & 0.0 & 9.0 & 100.0 & 0.0 \\
B2 & Sample from train distribution &  54.6 & 7.2 & 8.3 & 12.8 & 72.4 & 69.5 \\
B3 & Ordered uniform &  55.6 & 8.1 & 12.2 & 55.0 & 73.0 & 7.4 \\
\midrule
\multicolumn{5}{l}{\bf Supervised (\Section{sec:supervised_results})} \\
& \multicolumn{5}{l}{\bf Unstructured} \\
S1 & Discriminative linear &  71.0 & 36.0 & 31.6 & 30.7 & 73.3 & 27.1 \\
S2 & Discriminative MLP &   {\bf 75.9} & 30.4 & 27.7 & 41.1 & 82.8 & 13.0 \\
S3 & Gaussian mixture &   69.4 & 40.6 & 31.5 & 33.3 & 68.9 & 23.9 \\
\hdashline \\[-1em] 
& \multicolumn{5}{l}{\bf Structured} \\
S4 & \zhukov  & 75.2 & 18.1 & {\bf 45.4} & 54.4 & 90.7 & 7.4
\\
S5 & SMM, discriminative & 66.0 & 37.3 & 24.1 & 50.5 & 65.9 & 8.5 \\
S6 & SMM, generative & 60.5 & {\bf 49.4} & 28.7 & 46.6 & 52.4 & 10.6 \\
\midrule
\multicolumn{5}{l}{\bf Un- and Weakly-Supervised (\Section{sec:unsupervised_results})} \\
& \multicolumn{5}{l}{\bf Fully Unsupervised} \\
U1  & HSMM \small{(with opt.\ acc.\ assignment)} &  31.8 & 28.8 & 10.6 & 31.0 & 31.1 & 15.4 \\
\hdashline \\[-1em] 
& \multicolumn{5}{l}{\bf Ordering Supervision} \\
U2 & \richard &  40.8 & 14.0 & 12.1 & 55.0 & 49.8 & 7.4 \\
U3 & \zhukov ~ (\small without Narr.) &   69.5 & 0.2 & 2.8 & 55.0 & 97.2 & 7.4 \\
U4 & HSMM + Ord &  55.5 & 8.3 & 7.3 & 55.0 & 70.6 & 7.4 \\
\hdashline \\[-1em] 
& \multicolumn{5}{l}{\bf Narration Supervision} \\
U5    & HSMM + Narr & 65.7 & 9.6 & 8.5 & 35.1 & 84.6 & 4.5\\
\hdashline \\[-1em] 
& \multicolumn{5}{l}{\bf Ordering + Narration Supervision} \\ %
U6 & \zhukov &  {\bf 71.0} & 1.8 & {\bf 24.5} & 55.0 & 97.2 & 7.4 \\
U7 & HSMM + Narr + Ord &  61.2 & {\bf 15.9} & 17.2 & 55.0 & 73.7 & 7.4 \\
\bottomrule
\end{tabular}
}
\caption{
\label{tbl:full_results_table}
Model comparison on the CrossTask validation data. We evaluate primarily using all \acclabel accuracy and step \acclabel accuracy to evaluate action segmentation, and \steprecall to evaluate step recognition. 
}
\vspace{-0.2cm}
\end{table*}

\section{Results}

We first define several baselines based on dataset statistics (\Section{sec:dataset_statistic_baselines}), which we will find to be strong in comparison to past work. 
We then analyze each aspect of our proposed model on the dataset in a supervised training setting (\Section{sec:supervised_results}),
removing some error sources of unsupervised learning and evaluating whether a given model fits the dataset  \cite{liang2008analyzing}.
Finally,  we move to our main setting, the weakly-supervised setting of past work, incrementally adding step ordering and narration constraints
(see \Section{sec:supervision_conditions}) to evaluate the degree to which each helps (\Section{sec:unsupervised_results}).

\begin{figure}
\centering
\includegraphics[width=\linewidth]{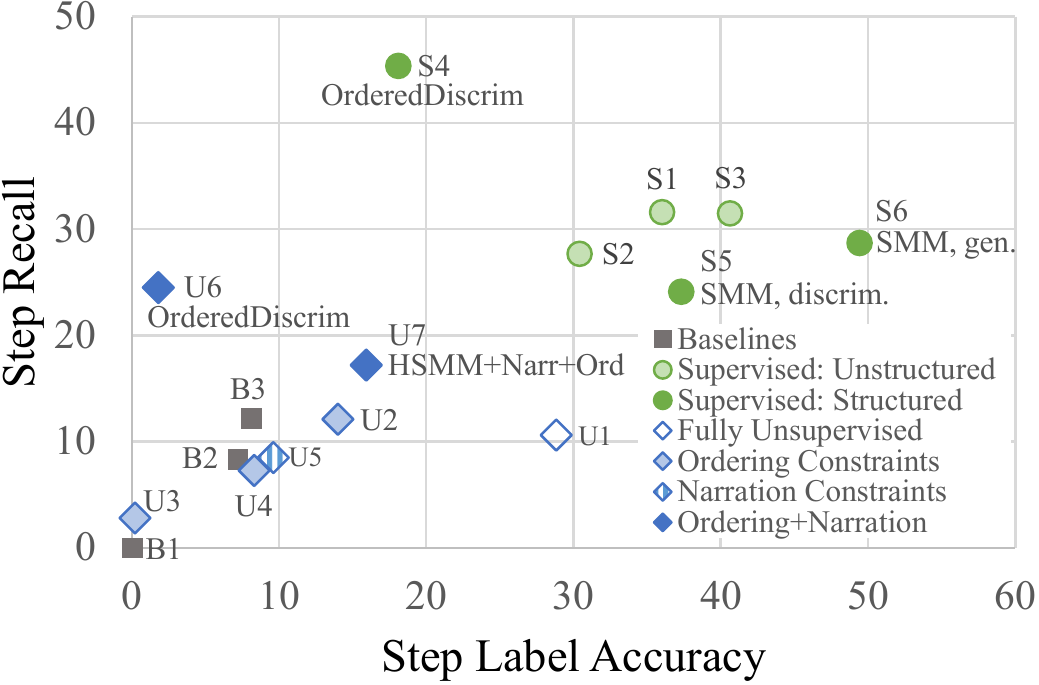}
\caption{\label{fig:results}Baseline and model performance on two key metrics: step label accuracy and step recall. Points are colored according to their supervision type, and labeled with their row number from \Table{tbl:full_results_table}. We also label particular important models.}
\vspace*{-1cm}
\end{figure}

Results are given in \Table{tbl:full_results_table} for models trained on the CrossTask training set of primary tasks, and evaluated on the held-out validation set. We will describe and analyze each set of results in turn. 
See \Figure{fig:results} for a plot of models' performance on two key metrics, and \Appendix{sec:a_segmentation_visualizations} for example predictions.

\subsection{Dataset Statistic Baselines}
\label{sec:dataset_statistic_baselines}

\Table{tbl:full_results_table} (top block) shows baselines that \emph{do not} use video (or narration) features, but predict steps according to overall statistics of the training data. These demonstrate characteristics of the data, and the importance of using multiple metrics.

\vspace{-0.2cm}
\paragraph{Predict background (B1)} %
Since most \timesteps %
are background, a model that predicts background \emph{everywhere} can obtain high overall \acclabel accuracy,
showing the importance of also using step \acclabel accuracy as a metric for action segmentation.

\vspace{-0.2cm}
\paragraph{Sample from the training distribution (B2)} For each \timestep in each video, we sample a label from the empirical distribution of step and background label frequencies for the video's task in the training data.

\vspace{-0.2cm}
\paragraph{Ordered uniform (B3)} 

\begin{figure*}
\begin{subfigure}[t]{0.45\textwidth}
\includegraphics[width=\linewidth]{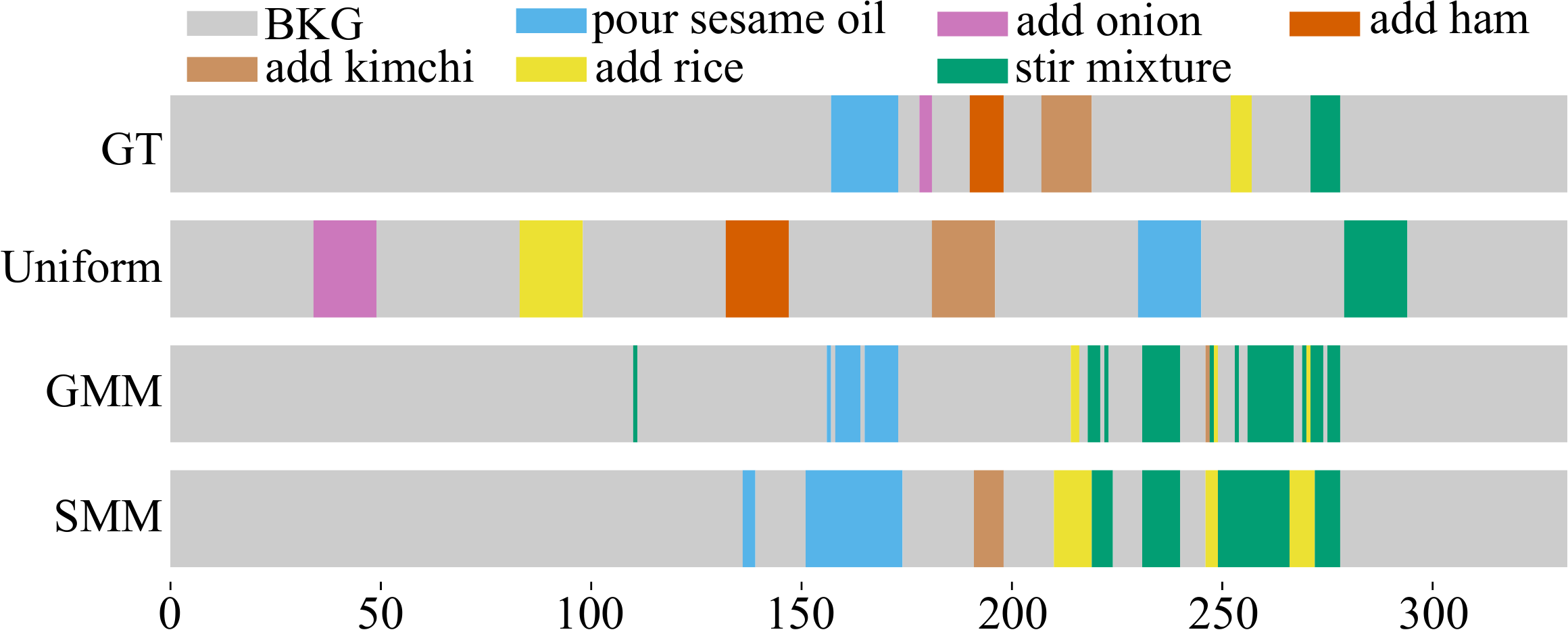}
\caption{\label{fig:supervised_segmentations_main}Step segmentations in the full supervision condition for a video from the \emph{make kimchi fried rice} task, comparing the ground truth (GT), ordered uniform baseline (Uniform), and predictions from the Gaussian mixture (GMM) and semi-Markov (SMM) models.}
\end{subfigure}
\hfill 
\begin{subfigure}[t]{0.51\textwidth}
\includegraphics[width=\linewidth]{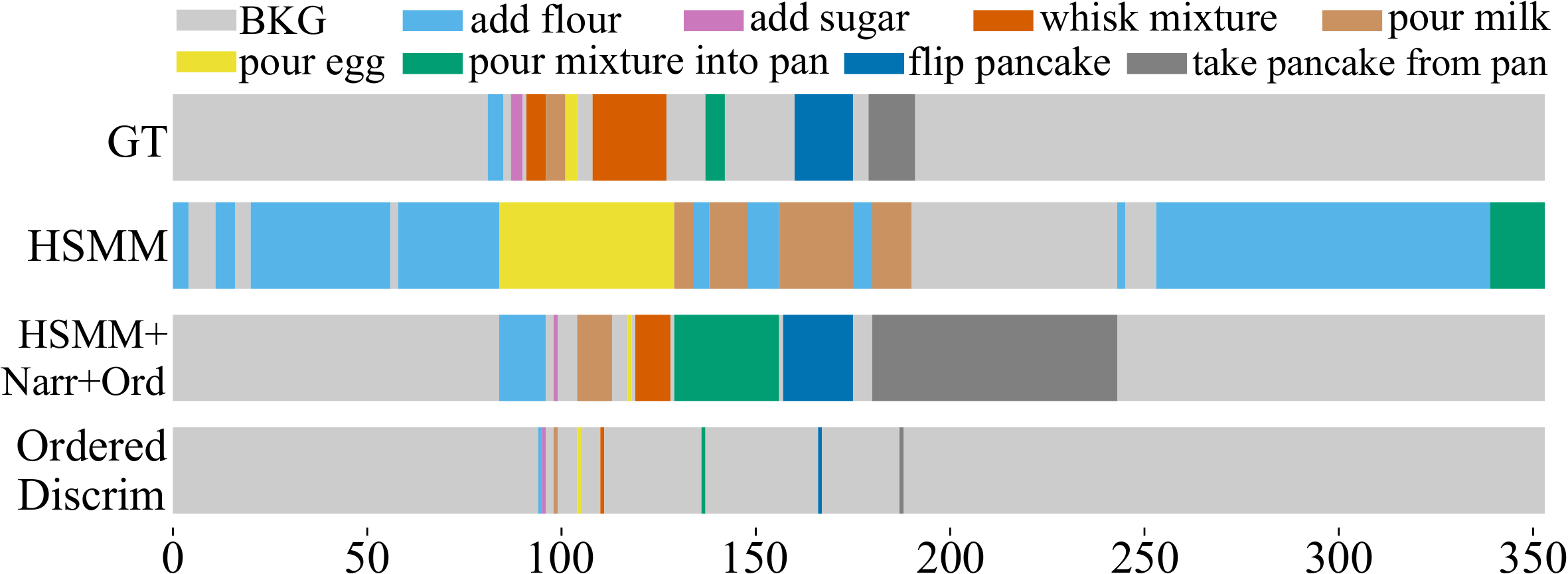}
\caption{\label{fig:unsupervised_segmentations_main}Step segmentations in the no- or weak-supervision conditions for a video from the \emph{make pancakes} task, comparing the ground truth (GT) to predictions from our model without (HSMM) and with constraint supervision (HSMM+Narr+Ord) and from \citet{zhukov2019cross} (\zhukov). 
}
\end{subfigure}
\caption{\label{fig:segmentations_main}Step segmentation visualizations for two sample videos in supervised (left) and unsupervised (right) conditions. The x-axes show timesteps, in seconds. 
See \Appendix{sec:a_segmentation_visualizations} for more visualizations.
}
\end{figure*}

For each video, we predict step regions in the canonical step order, separated by background regions. The length of each region is set so that all step regions in a video have equal duration, and the percentage of background timesteps is equal to the corpus average. See \emph{Uniform} in \Figure{fig:supervised_segmentations_main} for sample predictions.

Sampling each \timestep label independently from the task distribution (row B2), and using a uniform step assignment in the task's canonical ordering with background (B3) both obtain similar step \acclabel accuracy, but the ordered uniform baseline improves substantially on the \steprecall metric, indicating that step ordering is a useful inductive bias for \emph{step recognition}.

\subsection{Full Supervision}
\label{sec:supervised_results}

Models in the {\bf unstructured} block of \Table{tbl:full_results_table} are classification models applied independently to all \timesteps, allowing us to compare the performance of the feature models used as components in our structured models. 
We find that a Gaussian mixture model (row S3), which is used as the feature 
model in the HSMM, obtains comparable step recall and substantially higher step \acclabel accuracy than a discriminative linear classifer (row S1) similar to the one used in \citet{zhukov2019cross}, 
which is partially explained by the discriminative classifier overpredicting the background class (comparing Predicted Background \% for those two rows).
Using a higher capacity discriminative classifier, a neural net with a single hidden layer (MLP), improves performance over the linear model on several metrics (row S2);
however, the MLP still overpredicts background, substantially underperforming the Gaussian mixture on the step \acclabel accuracy metric.

In the {\bf structured} block of \Table{tbl:full_results_table}, we compare the full models which use step constraints \cite{zhukov2019cross} or learned transition distributions (the SMM) to model task structure.  
The structured models learn (or in the case of Zhukov et al., enforce) orderings over the steps, which greatly improve their sequence similarity scores when compared to the unstructured models, and decrease step fragmentation (as measured by num.\ segments). \Figure{fig:supervised_segmentations_main} shows predictions for a typical video, demonstrating this decreased fragmentation.\footnote{We also perform an ablation study to understand the effect of the duration model. See \Appendix{sec:hmm_ablation} for details.}

We see two trends in the supervised results:

(1) Generative models obtain substantially higher step \acclabel accuracy than discriminative models of the same or similar class. 
This is likely due to the fact that the 
generative models directly parameterize the step distribution. %
(See \Appendix{sec:gen_or_dis}.)

(2) Structured sequence modeling naturally improves performance on sequence-level metrics (sequence similarity and number of segments predicted) over the unstructured models. However, none of the learned structured models improve on the strong ordered uniform baseline (B3) which just predicts the canonical ordering of a task's steps (interspersed with background regions). This will motivate using this canonical ordering as a constraint in unsupervised learning.

Overall, the SMM models obtain strong action segmentation performance (high step label accuracy without fragmenting segments or overpredicting background).

\subsection{No or Weak Supervision}
\label{sec:unsupervised_results}

Here models are trained without supervision for the labels $l_{1:T}$. 
We compare models trained without any constraints, to those that use constraints from step ordering and narration, %
in the {\bf Un- and Weakly Supervised} block of \Table{tbl:full_results_table}. Example outputs are shown in \Appendix{sec:a_segmentation_visualizations}.

Our generative HSMM model affords training without any constraints (row U1).
This model has high step \acclabel accuracy (compared to the other unsupervised models) but low all \acclabel accuracy, and similar scores for both metrics. This hints, and other metrics confirm, that the model is not adequately distinguishing steps from background:
the percentage of predicted background is very low (31\%) compared to the ground truth (72\%, row GT). 
See \emph{HSMM} in \Figure{fig:unsupervised_segmentations_main} for predictions for a typical video.
These results are attributable to features within a given video (even across step types)  being more similar than features of the same step type in different videos 
(see \Appendix{sec:a_feature_visualizations} for feature visualizations).
The induced latent model states typically capture this inter-video diversity, rather than distinguishing steps across tasks. %

We next add in constraints from the canonical step {\bf ordering}, which our supervised results showed to be a strong inductive bias.  
Unlike in the fully unsupervised setting, the HSMM model with ordering (HSMM+Ord, row U4) \emph{learns} to distinguish steps from background when constrained to predict each step region once in a video, with predicted background \timesteps (70.6\%) close to the ground-truth (72\%). 
However, performance of this model is still very low on the task metrics -- comparable to or underperforming the ordered uniform baseline with background (row B3) on all metrics.

This constrained step ordering setting also allows us to apply \richard \citep{richard2018action} and \zhukov \citep{zhukov2019cross}. \richard obtains high step \acclabel accuracy, but substantially underpredicts background, as evidenced by both the all \acclabel accuracy and the low predicted background percentage.
The tendency of \zhukov to overpredict background which we saw in the supervised setting (row S4) is even more pronounced in this weakly-supervised setting (row U3), resulting in scores very close to the predict background baseline (B1).

Next, we use {\bf narration} constraints (U5), which are enforced only during training time, following \citet{zhukov2019cross}. Narration constraints substantially improve all label accuracy (comparing U1 and U5). However, the model overpredicts background, likely because it doesn't enforce each step type to occur in a given video. Overpredicting background causes step \acclabel accuracy and \steprecall to decrease.

Finally, we compare the HSMM and \zhukov models when using both narration constraints (in training) and ordering constraints (in training and testing) in the {\bf ordering + narration} block. Both models benefit substantially from narration on all metrics when compared to using only ordering supervision, more than doubling their performance on step \acclabel accuracy and \steprecall (comparing U6 and U7 to U3 and U4).

Our weakly-supervised results show that:

(1) \emph{Both} action segmentation metrics -- all \acclabel accuracy and step \acclabel accuracy -- are important to evaluate whether models adequately distinguish meaningful actions from background.

(2) Step constraints derived from the canonical step ordering provide a strong inductive bias for unsupervised step induction. Past work requires these constraints and the HSMM, when trained without them, does poorly, learning to capture diversity across videos rather than to identify steps. 

(3) However, ordering supervision alone is not sufficient to allow these models to learn better segmentations than a simple baseline that just uses the ordering to assign labels (\emph{ordered uniform}); narration is also required. %

\subsection{Comparison to Past Work}

\begin{table}[t]
\centering
\small
\begin{tabular}{lccc}
\toprule
& All \Acclabel & Step \Acclabel  & Step \\
& Acc. & Acc. & \Recovery \\
\midrule
    \zhukov & 71.3 & 1.2 & 17.9  \\
    HSMM+Narr+Or & 66.0 & 5.6 & 14.2 \\
\bottomrule
\end{tabular}
\caption{\label{tbl:predict_background}Unsupervised and weakly supervised results in the cross-validation setting.
}
\vspace{-0.4cm}
\end{table}

Finally, 
we compare our full model to the \zhukov model of \citet{zhukov2019cross} in the primary data evaluation setup from that work: averaging results over 20 random splits  of the primary data (\Table{tbl:predict_background}).
This is a low data setting which uses only 30 videos per task as training data in each split.

Accordingly, both models have lower performance,
although the relative ordering is the same: higher step \acclabel accuracy for the HSMM, and higher all \acclabel accuracy and \steprecall for \zhukov. Although in this low-data setting, 
models overpredict background even more, this problem is less pronounced for the HSMM: 97.4\% of timesteps for \zhukov are predicted background (explaining its high all \acclabel accuracy), and 87.1\% for HSMM.

\vspace{-0.5em}
\section{Discussion}
\vspace{-0.5em}

We find that unsupervised action segmentation in naturalistic instructional videos is greatly aided by the inductive bias given by typical step orderings within a task, and narrative language 
describing the actions being done.
While some results are more mixed (with the same supervision, different models are better on different metrics), we do observe that
across settings and metrics, step ordering and narration increase performance. 
Our results also illustrate the importance of strong baselines: without weak supervision from step orderings and narrative language, even state-of-the-art unsupervised action segmentation models operating on rich video features underperform feature-agnostic baselines.
We hope that future work will continue to evaluate broadly.

While action segmentation in videos from diverse domains remains challenging -- videos contain both a large variety of types of depicted actions, and high visual variety in how the actions are portrayed -- we find that structured generative models provide a strong benchmark for the task due to their abilities to capture the full diversity of action types (by directly modeling distributions over action occurrences), and to benefit from weak supervision. 
Future work might explore methods for incorporating richer learned representations both of the diverse visual observations in videos, and their narrative descriptions, into such models.

\section*{Acknowledgments}
\vspace{-0.8em}
Thanks to Dan Klein, Andrew Zisserman, Lisa Anne Hendricks, Aishwarya Agrawal, G\'abor Melis, Angeliki Lazaridou, Anna Rohrbach, Justin Chiu, Susie Young, the DeepMind language team, and the anonymous reviewers for helpful feedback on this work.
DF is supported by a Google PhD Fellowship.

\bibliography{refs}
\bibliographystyle{acl_natbib}

\clearpage

\appendix

\section{Optimization}
\label{sec:a_optimization}

For both training conditions 
for our semi-Markov models 
that require gradient descent
(generative unsupervised and discriminative supervised),
we initialize parameters randomly and use Adam \citep{kingma2014adam} with an initial learning rate of 5e-3, a batch size of 5 videos, and decay the learning rate when training log likelihood does not decrease for more than one epoch. %

\section{Features}
\label{sec:a_feature_models}

For our features $x_{1:T}$, we use the same base features as \citet{zhukov2019cross}. There are three feature types: activity recognition features, produced by an I3D model \citep{carreira2017quo} trained on the Kinetics-400 dataset \cite{kay2017kinetics}; object classification features, from a ResNet-152 \cite{he2016deep} trained on ImageNet \citep{imagenet15}, and audio classification features\footnote{\url{https://github.com/tensorflow/models/tree/master/research/audioset/vggish}} from the VGG model \cite{simonyan2015vgg} trained by \citet{hershey2017cnn} on a preliminary version of the YouTube-8M dataset \cite{abu2016youtube}.\footnote{We also experiment with using features derived from transcribed narration in \Appendix{sec:narration_features}.}

For the generative mdoels which use Gaussian emission distributions, we apply PCA to the base features above to reduce the feature dimensionality and decorrelate dimensions.
We perform PCA separately for features within task and within each feature group (I3D, ResNet, and audio features), but on features from all videos within that task. We use 100 components for each feature group, which explained roughly 70-100\% of the variance in the features, depending on the task and feature group. The 100-dimensional PCA representations for the I3D, ResNet, and audio features for each frame, at timestep $t$, are then concatenated to give a 300-dimensional vector for the frame, $x_t$.

\section{Unsupervised Evaluation}
\label{sec:a_unsupervised_evaluation}
The HSMM model, when trained in a fully unsupervised setting, induces class labels for regions in the video; however while these class labels are distinct, they do not correspond \emph{a priori} to any of the actual region labels (which can be step types, or background) for our task. Just as with other unsupervised tasks and models (\eg, part-of-speech induction), we need a mapping from these classes to step types (and background) in order to evaluate the model's predictions.
We follow the evaluation procedure of past work \cite{sener2018unsupervised,Sener15unsupervised} by finding the mapping from model states to region labels that maximizes label accuracy, averaged across all videos in the task, using the Hungarian method \cite{Kuhn55thehungarian}.
This evaluation condition is only used in the ``Unsupervised'' section of \Table{tbl:full_results_table} (in the rows marked with \emph{optimal accuracy assignment}).

\section{Evaluation Metrics}
\label{sec:a_evaluation_metrics}

\paragraph{Label accuracy} 
The standard metric for action segmentation \cite{sener2018unsupervised,richard2018action} 
is timestep label accuracy, in datasets with a large amount of background, label accuracy on non-background timesteps. 
The CrossTask dataset has multiple reference step labels in the groundtruth for around 1\% of timesteps, due to noisy region annotations that overlap slightly.
We obtain a single reference label for these timesteps by taking the step that appears first in the canonical step ordering for the task. We then compute accuracy of the model predictions against these reference labels across all \timesteps and all videos for a task (in the \emph{all \acclabel accuracy} condition), or by filtering to those \timesteps which have a step label (non-background) in the reference (to focus on the model's ability to accurately predict step labels), in the \emph{step \acclabel accuracy} condition.

\paragraph{Step recall} This metric \cite{zhukov2019cross}
measures a model's ability to pick out instants for each of the possible step types for a task, if they occur in a video. 
The model of \citet{zhukov2019cross} predicted a single frame for each step type; while our extension of their model, \zhukov, and our HSMM model can predict multiple, when computing this metric we obtain a single frame for each step type to make the numbers comparable to theirs.%
When a model predicts multiple frames per step type, we obtain a single one by taking the one closest to the middle of the temporal extent of the predicted frames for that step type.
We then apply their recall metric:
First, count the number of \emph{recovered steps}, step types from the true labels for the video that were identified by one of the predicted labels (have a predicted label of the same type at one of the true label's frames). These recovered step counts are summed across videos in the evaluation set for a given task, and normalized by the maximum number of possible recovered steps (the number of step types in each video, summed across videos) to produce a step recall fraction for the task.

\paragraph{Sequence similarity} This measures the similarity of the predicted sequence of regions in a video against the true sequence of regions. As in speech recognition, we are interested in the high-level sequence of steps recognized in a video (and wish to abstract away from noise in the boundaries of the annotated regions). 
We first compute the negated Levenshtein distance between the true sequence of steps and background $r_1, \ldots, r_K$ for a video and the and predicted sequence $\hat r_1, \ldots, \hat r_K'$. The negated distance for the sequence pairs for a given video are scaled to be between 0 and 100, where 0 indicates the Levenshtein distance is the maximum possible between two sequences of their respective lengths, and 100 corresponds to the sequences being identical. These similarities are then averaged across all videos in a task.

\section{Comparing Generative and Discriminative Models}
\label{sec:gen_or_dis}

We observe that the generative models tend to obtain higher performance on the action segmentation task, as measured by step \acclabel accuracy, than discriminative models of the same or similar class.
We attribute this finding to two factors: first, the generative models explicitly parameterize probabilities for the steps, allowing better modeling of the full distribution of step labels.
Second, the discriminative models are trained to optimize $p(l_t \mid x_t)$ for all \timesteps $t$.
We would expect that this would produce better accuracies on metrics aligned with this objective \cite{klein2002conditional} -- and indeed the all \timestep accuracy is higher for the discriminative models.
However, the discriminative models' high accuracy often comes at the expense of predicting background more frequently, leading to lower performance on step \acclabel accuracy.

\section{Duration Model Ablation}
\label{sec:hmm_ablation}

\begin{table}[h]
\centering
\small
\scalebox{0.91}{
\begin{tabular}{lcccc}
\toprule
      & All \Acclabel & Step \Acclabel & Step      & Seq.   \\
Model & Acc.          & Acc.           & \Recovery & Sim. \\
\midrule
\multicolumn{5}{l}{\bf Supervised} \\
SMM, gen. &  60.5 & 49.4 & 28.7 & 46.6 \\
MM, gen. &   60.1 & 48.6 & 28.2 & 46.8 \\
SMM, disc. & 66.0 & 37.3 & 24.1 & 50.5 \\
MM, disc. & 62.8 & 32.2 & 20.1 & 41.8 \\
\midrule
\multicolumn{5}{l}{\bf Weakly-Supervised} \\
HSMM &  31.8 & 28.8 & 10.6 & 31.0 \\
HMM &   28.8 & 30.8 & 10.3 & 29.9 \\
HSMM+Ord+Narr &  61.2 & 15.9 & 17.2 & 55.0 \\
HMM+Ord+Narr & 60.6 & 17.0 & 20.0 & 55.0 \\
\bottomrule
\end{tabular}
}
\caption{\label{tbl:hmm}Comparison between the semi-Markov and hidden semi-Markov models (SMM and HSMM) with the Markov and hidden Markov (MM and HMM) models, which ablate the semi-Markov's duration model.}
\end{table}

We examine the effect of the (hidden) semi-Markov model's Poisson duration model by comparing to a (hidden) Markov model (HMM in the unsupervised/weakly-supervised settings, or MM in the supervised setting). 
We use the model as described in \Section{sec:structured_model} except for fixing all durations to be a single timestep. We then train as described in \Section{sec:training}.
While this does away with explicit modeling of duration, the transition distribution still allows the model to learn expected durations for each region type by implicitly parameterizing a geometric distribution over region length.
Results are shown in \ref{tbl:hmm}.
We observe that results are overall very similar, with the exceptions that removing the duration model decreases performance substantially on all metrics in the discriminative supervised setting, and increases performance on step label accuracy and step recall in the constrained unsupervised setting (HSMM+Ord+Narr and HMM+Ord+Narr). 
This suggests 
that the HMM transition distribution is able to model region duration as well as the HSMM's explicit duration model,
or that duration overall plays a small role in modeling in most settings relative to the importance of the features.

\begin{figure*}[t!]
\centering
\begin{subfigure}[b]{.40\linewidth}
\includegraphics[width=\linewidth]{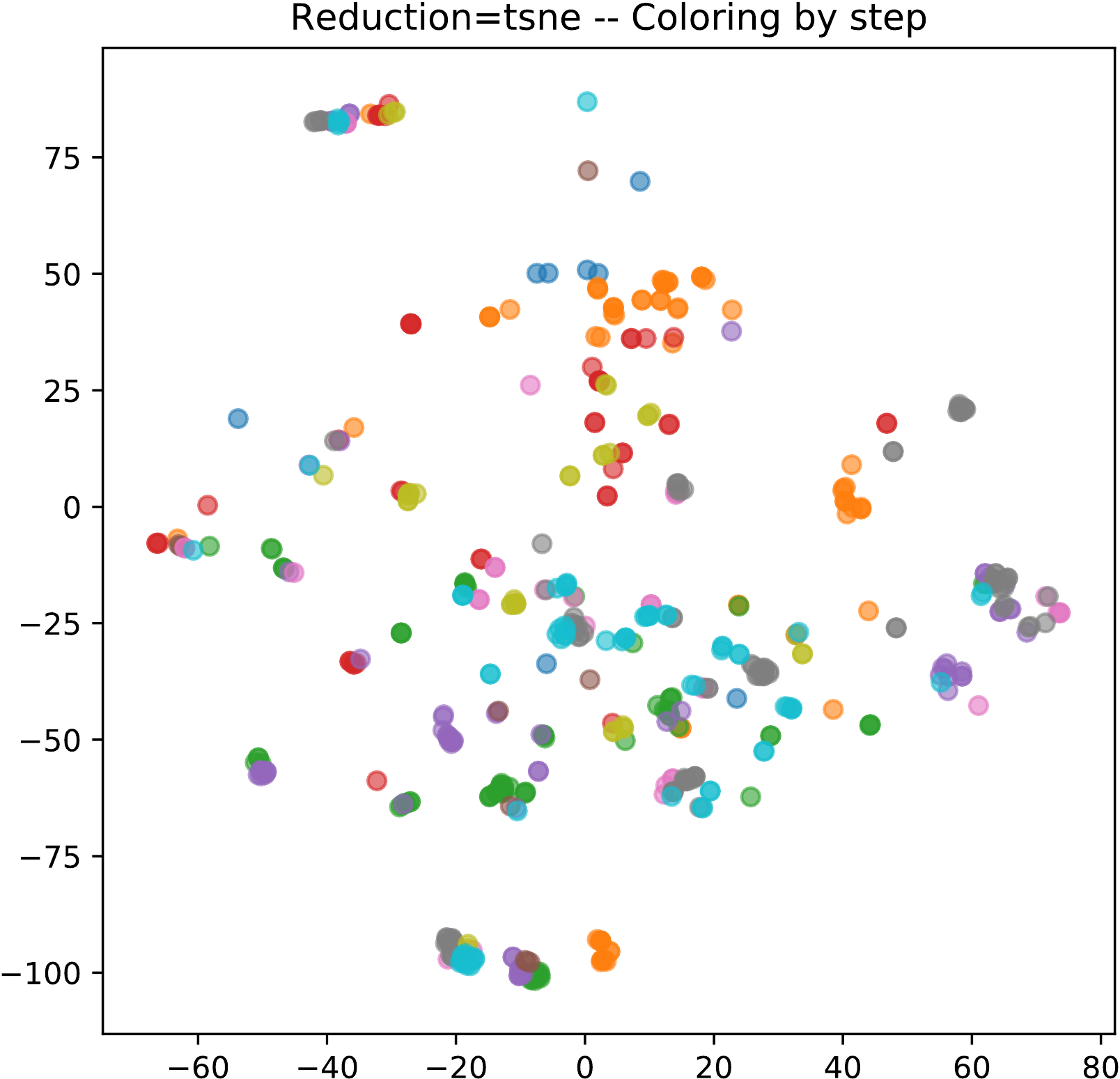}
\caption{\label{fig:feats_by_step}Feature vectors colored by their step label in the reference annotations.}
\end{subfigure}
\hspace{2em}
\begin{subfigure}[b]{.40\linewidth}
\includegraphics[width=\linewidth]{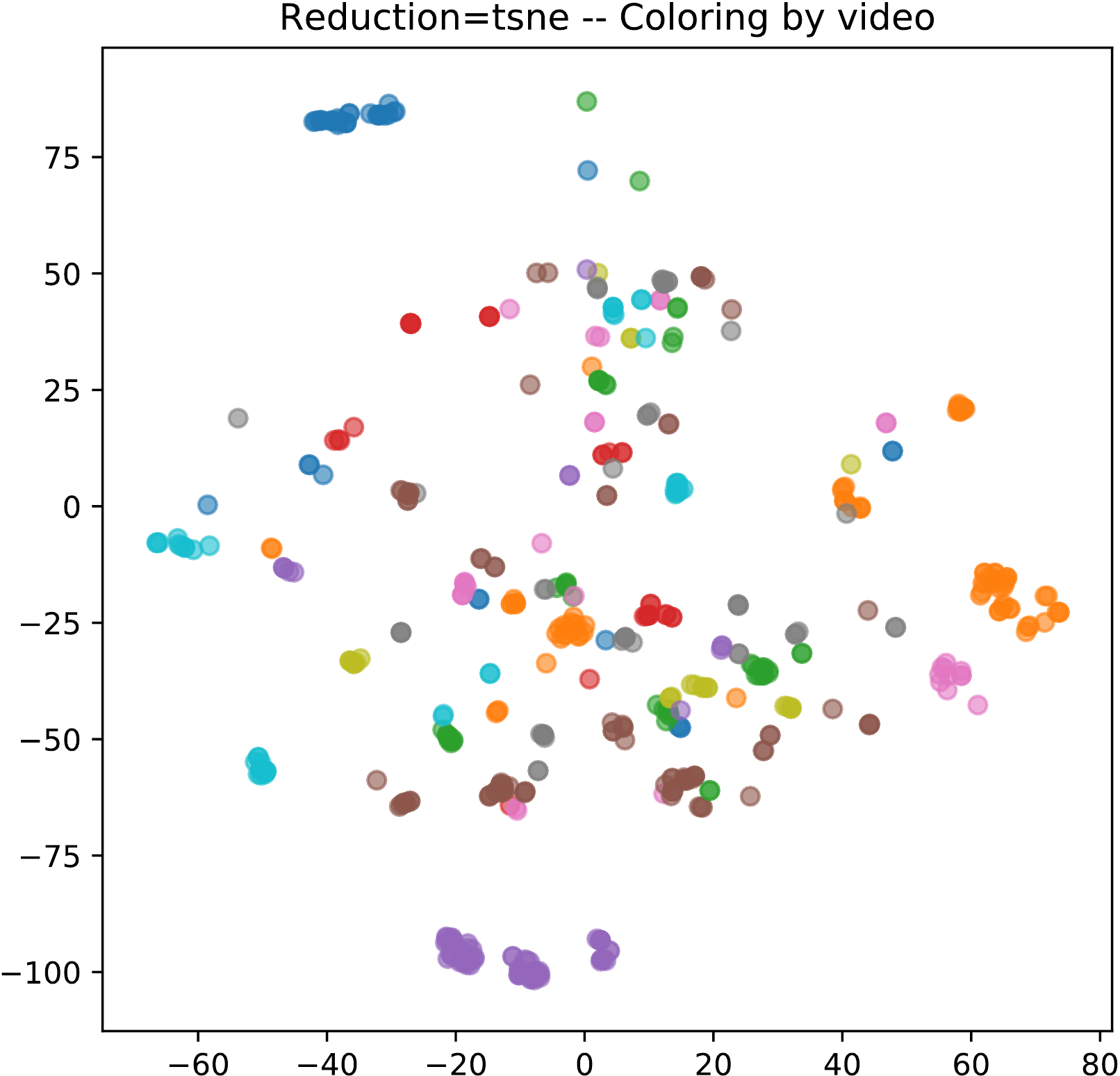}
\caption{\label{fig:feats_by_video}Feature vectors colored by the id of the video they occur in.}

\end{subfigure}
\caption{t-SNE plots of feature vectors for 20 videos from the \emph{change a tire} task, showing feature similarity.}
\end{figure*}

\section{Narration Features}
\label{sec:narration_features}
The benefit of narration-derived hard constraints on labels (following past work by \citealt{zhukov2019cross}) raises the question of how much narration would help when used to provide features for the models. We obtain narration features for each video using FastText word embeddings \cite{mikolov2018advances} for the video's time-aligned transcribed narration (see \citealt{zhukov2019cross} for details on this transcription), pooled within a sliding window to allow for imperfect alignment between activities mentioned in the narration and their occurrence in the video.
The features for a given timestep $t$ are produced by a weighted sum of embeddings for all the words in the transcribed narration within a 5-second window of $t$ (\ie from $t-2$ to $t+2$), weighted using a Hanning window\footnote{\url{https://docs.scipy.org/doc/numpy/reference/generated/numpy.hanning.html}} (so that words in the center of each window are most heavily weighted for that window). We did not tune the window size, or experiment with other weighting functions. 
The word embeddings are pretrained on Common Crawl, and are not fine-tuned with the rest of the model parameters. 

Once these narration features are produced, as above, we treat them in the same way as the other feature types (activity recognition, object classification, and audio) described in \Appendix{sec:a_feature_models}: reducing their dimensionality with PCA, and concatenating them with the other feature groups to produce the features $x_t$.

In Table~\ref{tbl:narration_features}, we show performance of key supervised and weakly-supervised models on the validation set, when using these narration features in addition to activity recognition, object detection, and audio features. Narration features improve performance over the corresponding systems from Table~\ref{tbl:full_results_table} (differences are shown in parentheses) in 13 out of 15 cases, typically by 1-4\%. 

\begin{table}[t]
\centering
\small
\scalebox{0.92}{
\begin{tabular}{lccc}
\toprule
& All \Acclabel & Step \Acclabel  & Step \\
& Acc. & Acc. & \Recovery \\
\midrule
    \multicolumn{3}{l}{\bf Supervised} \\
    Gaussian mixture & 70.4 (+1.0) & 43.7 (+3.1) & 34.9 (+3.4) \\
SMM, generative & 63.3 (+2.8) & 53.2 (+3.8) & 32.1 (+3.4) \\
\midrule
 \multicolumn{3}{l}{\bf Weakly-Supervised} \\
HSMM+Ord& 53.6 (-1.9) & \phantom{0}9.5 (+1.2) & \phantom{0}8.5 (+1.2) \\
HSMM+Narr& 68.9 (+3.2) & \phantom{0}8.0 (-1.6) & 12.6 (+4.1) \\
HSMM+Narr+Ord & 64.3 (+3.1) & 17.9 (+2.0) & 21.9 (+4.7) \\
\bottomrule
\end{tabular}
}
\caption{\label{tbl:narration_features}Performance of key supervised and weakly-supervised models on the validation data when adding narration vectors as features. Numbers in parentheses give the change from adding narration vectors to the systems from Table~\ref{tbl:full_results_table}.}
\end{table}

\section{Feature Visualizations}
\label{sec:a_feature_visualizations}

To give a sense for feature similarities both within step types and within a video, we visualize feature vectors for 20 videos randomly chosen from the \emph{change a tire} task, dimensionality-reduced using t-SNE \cite{maaten2008visualizing} so that similar feature vectors are close in the visualization. 

\Figure{fig:feats_by_step} shows feature vectors colored by step type: we see little consistent clustering of feature vectors by step. On the other hand, we observe a great deal of similarity across step types within a video (see \Figure{fig:feats_by_video}); when we color feature vectors by video, different steps from the \emph{same} video are close to each other in space. These together suggest that better featurization of videos can improve action segmentation.

\section{Segmentation Visualizations}
\label{sec:a_segmentation_visualizations}

\begin{figure*}
\centering
\includegraphics[width=\linewidth]{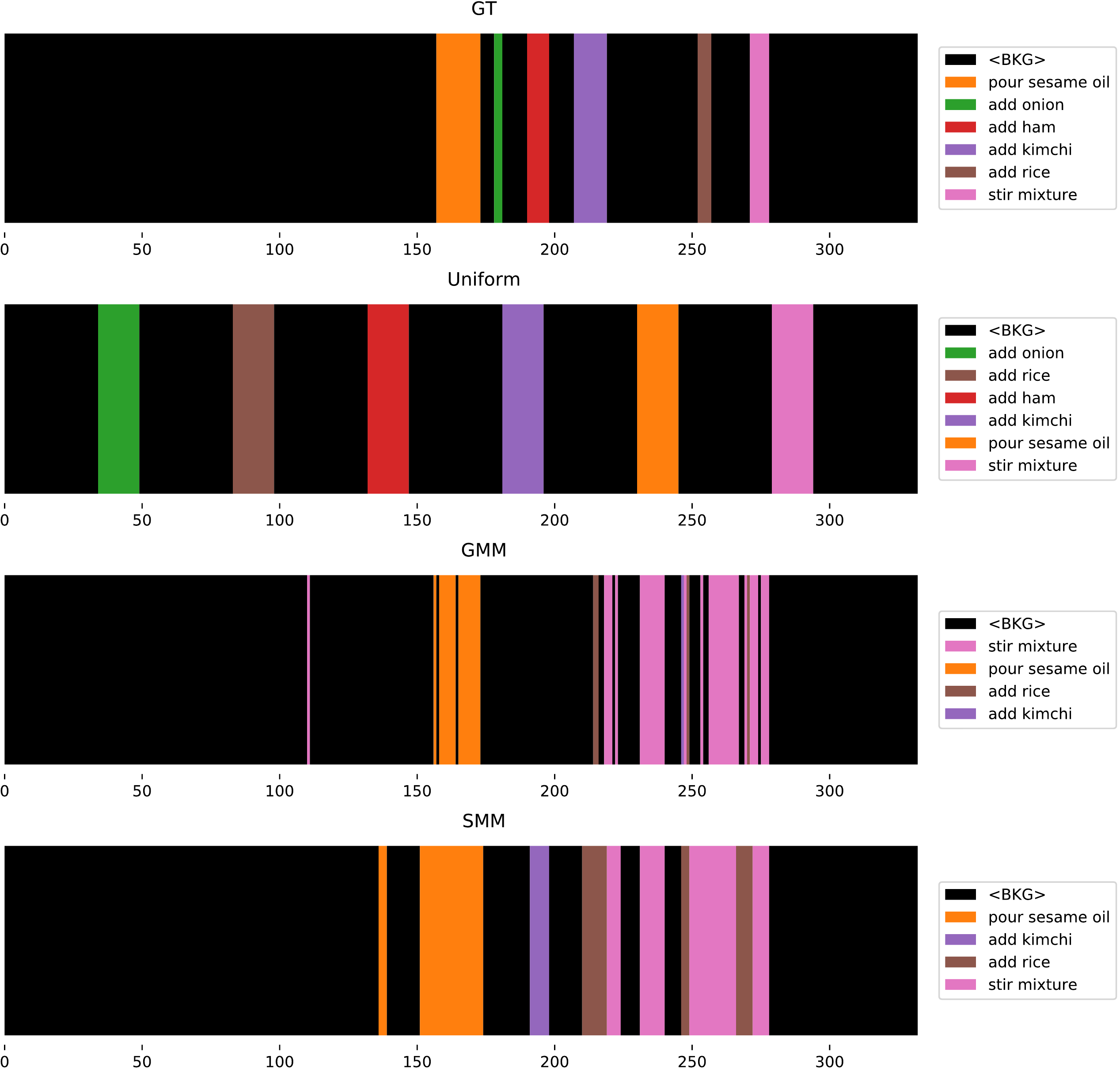}
\caption{\label{fig:supervised_1}{\bf Supervised segmentations} We visualize segmentations from the validation set for a video from the task \emph{make kimchi fried rice}. We show the ground truth (GT), ordered uniform baseline (Uniform), and predictions from the unstructured Gaussian mixture model (GMM), and structured semi-Markov model (SMM) trained in the supervised setting. Predictions from the unstructured model are more fragmented than predictions from the SMM. The x-axis gives the timestep in the video.}
\end{figure*}

\begin{figure*}
\centering
\includegraphics[width=\linewidth]{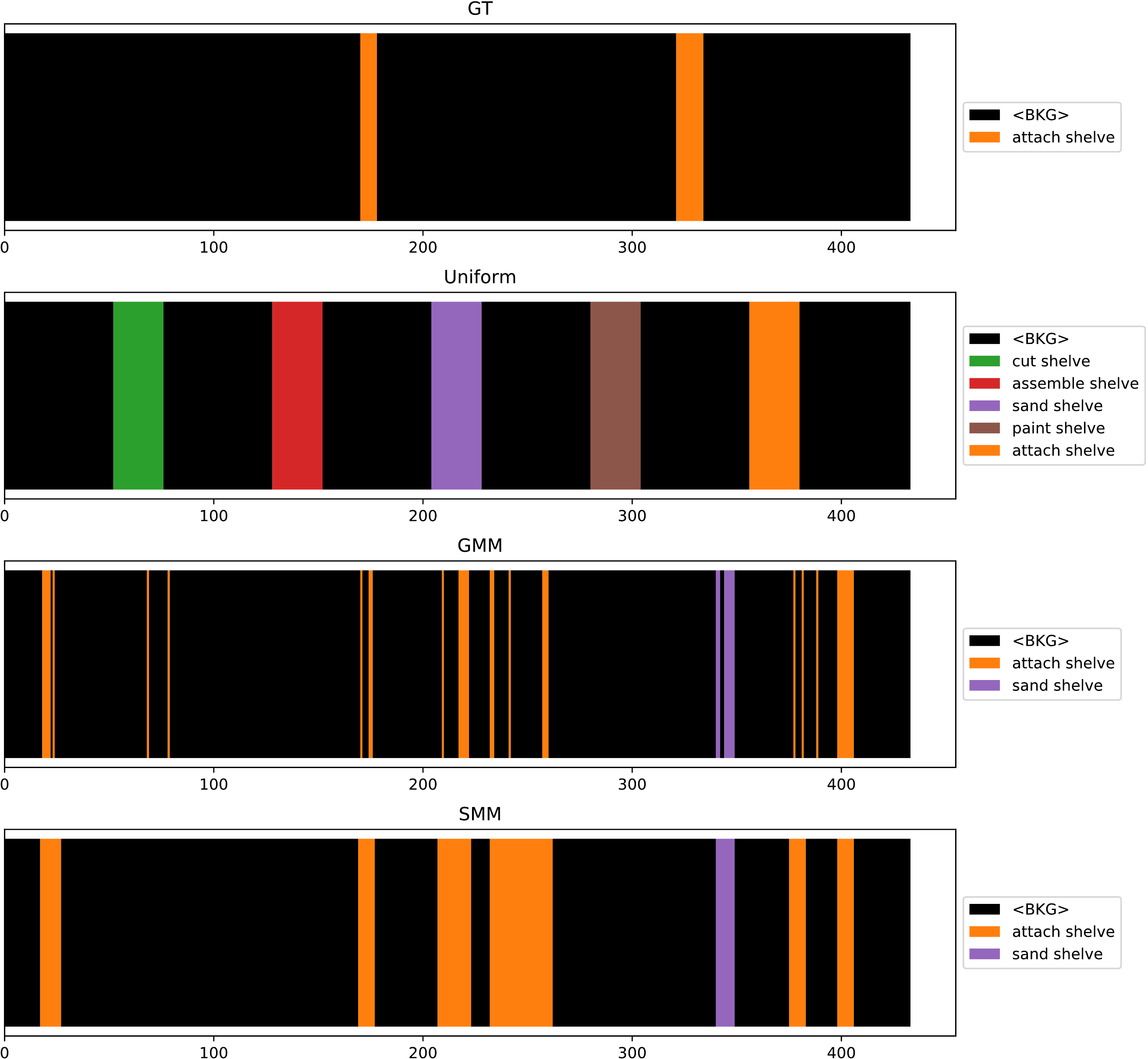}
\caption{\label{fig:supervised_2}{\bf Supervised segmentations} We visualize segmentations from the validation set for a video from the task \emph{build simple floating shelves}. We show the ground truth (GT), ordered uniform baseline (Uniform), and predictions from the unstructured Gaussian mixture model (GMM), and structured semi-Markov model (SMM) trained in the supervised setting. Predictions from the unstructured model are more fragmented than predictions from the SMM. The x-axis gives the timestep in the video.}
\end{figure*}

In the following pages, we show example segmentations from the various systems. \Figure{fig:supervised_1} and \ref{fig:supervised_2} visualize predicted model segmentations for the unstructured Gaussian mixture and structured semi-Markov model in the supervised setting, in comparison to the ground-truth and the ordered uniform baseline.
We see that while both models typically make similar predictions in the same temporal regions of the video, the structured model produces steps that are much less fragmented.

\begin{figure*}
\centering
\includegraphics[width=\linewidth]{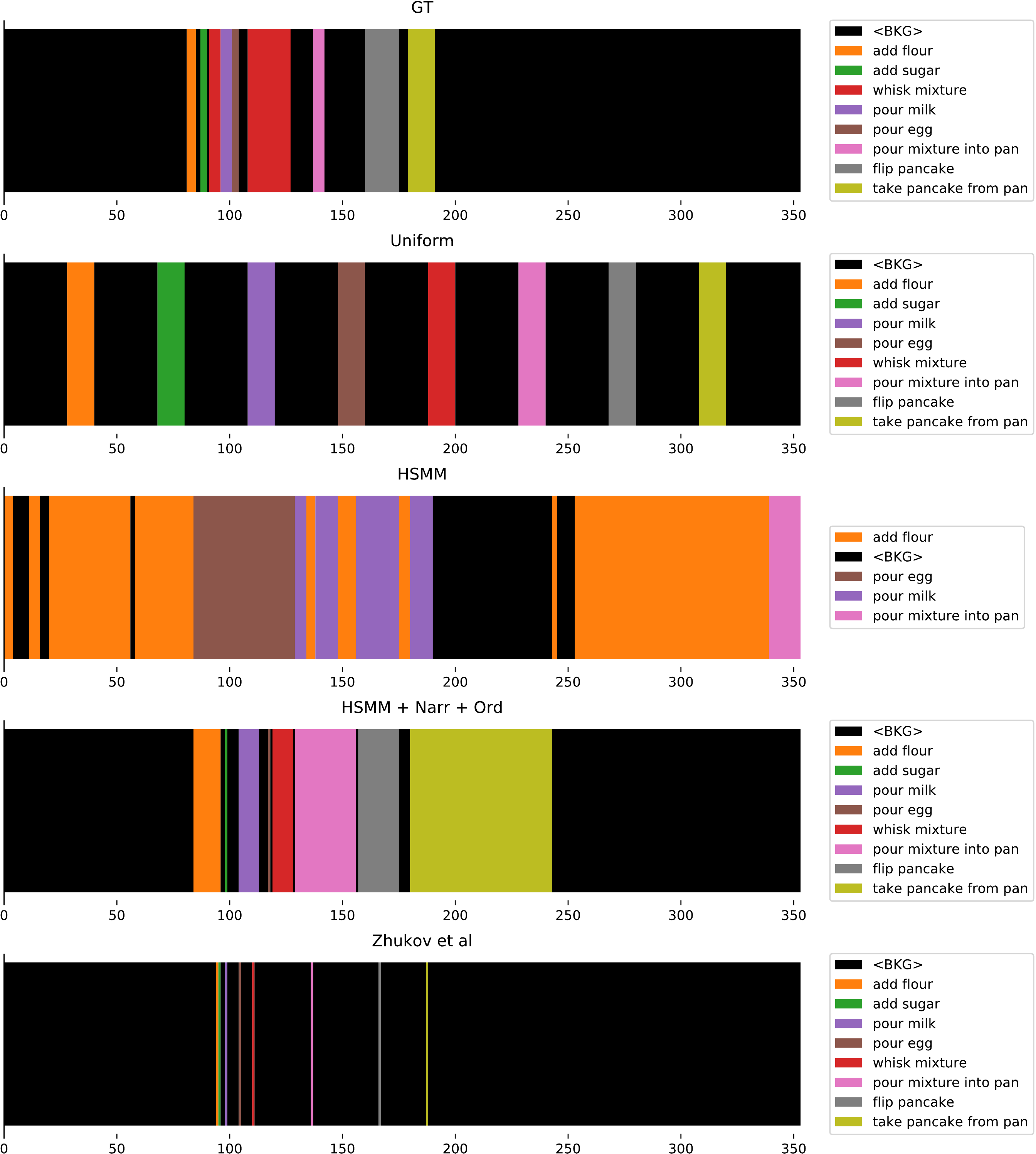}
\caption{\label{fig:unsupervised_1}{\bf Unsupervised and weakly-supervised segmentations} We visualize segmentations from the validation set for a video from the task \emph{make pancakes}. We show the ground truth (GT), ordered uniform baseline (Uniform), and predictions from the hidden semi-markov trained without constraints (HSMM) and with constraints from narration and ordering (HSMM+Narr+Ord), and the system of Zhukov et al. The x-axis gives the timestep in the video.}
\end{figure*}

\begin{figure*}
\centering
\includegraphics[width=\linewidth]{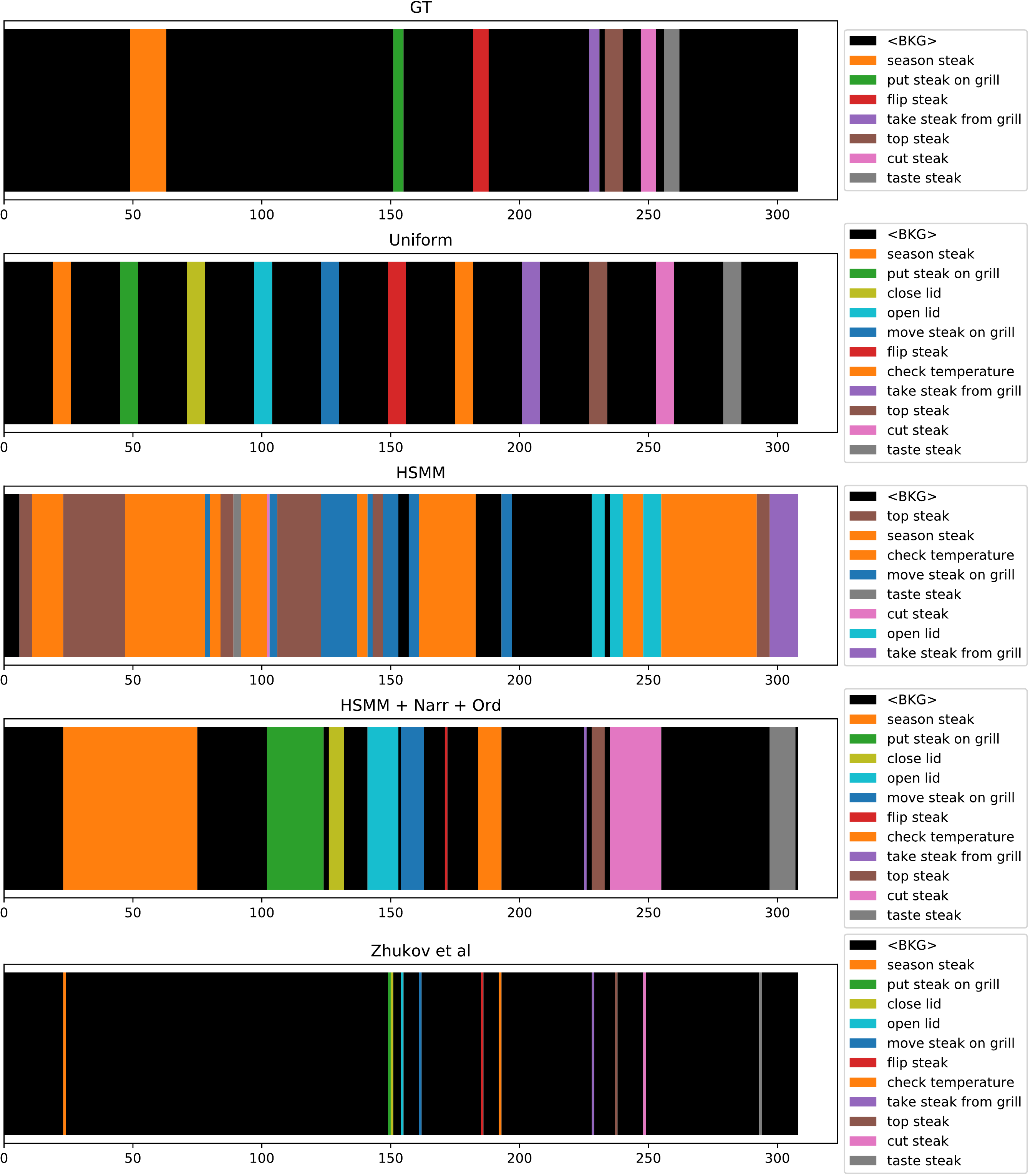}
\caption{\label{fig:unsupervised_2}{\bf Unsupervised and weakly-supervised segmentations} We visualize segmentations from the validation set for a video from the task \emph{grill steak}. We show the ground truth (GT), ordered uniform baseline (Uniform), and predictions from the hidden semi-markov trained without constraints (HSMM) and with constraints from narration and ordering (HSMM+Narr+Ord), and the system of Zhukov et al. The x-axis gives the timestep in the video.}
\end{figure*}

\Figure{fig:unsupervised_1} and \ref{fig:unsupervised_2} visualize segmentations in the unsupervised and weakly-supervised settings for the HSMM model and \zhukov of \citet{zhukov2019cross}.
The unsupervised HSMM has difficulty distinguishing steps from background (see \Appendix{sec:a_feature_visualizations}), while the model trained with weak supervision from ordering and narration (HSMM+Ord+Narr) is better able to induce meaningful steps. The \zhukov model, although it has been modified to allow predicting multiple timesteps per step, collapses to predicting a single label, background, nearly everywhere, which we conjecture is because the model is discriminatively trained: jointly inferring labels that are easy to predict, and the model parameters to predict them.

\end{document}